\documentclass{article}

\usepackage{microtype}
\usepackage{graphicx}
\usepackage{subfigure}
\usepackage{booktabs} 
\usepackage{array}
\usepackage{amsmath,amsfonts,bm} 

\usepackage{hyperref}
\usepackage{pifont}

\newcommand{\norm}[1]{\left\lVert#1\right\rVert}
\newcommand{\cmark}{\ding{51}}%
\newcommand{\xmark}{\ding{55}}%

\usepackage[accepted]{icml2021}

\icmltitlerunning{Whitening for Self-Supervised Representation Learning}

\begin{document}

\twocolumn[
\icmltitle{Whitening for Self-Supervised Representation Learning}



\icmlsetsymbol{equal}{*}

\begin{icmlauthorlist}
\icmlauthor{Aleksandr Ermolov}{unitn}
\icmlauthor{Aliaksandr Siarohin}{unitn}
\icmlauthor{Enver Sangineto}{unitn}
\icmlauthor{Nicu Sebe}{unitn}
\end{icmlauthorlist}

\icmlaffiliation{unitn}{Department of Information Engineering and Computer Science (DISI), University of Trento, Italy}

\icmlcorrespondingauthor{Aleksandr Ermolov}{aleksandr.ermolov@unitn.it}

\icmlkeywords{Self-Supervised Learning; Unsupervised Learning; Representation Learning; Contrastive loss; Whitening}

\vskip 0.3in
]



\printAffiliationsAndNotice{}  

\begin{abstract}
Most of the current self-supervised representation learning (SSL) methods are based on the {\em contrastive loss} and the instance-discrimination task, where augmented versions of the same image instance  (``positives'') are contrasted with instances extracted from other images (``negatives''). For the learning to be effective, many negatives should be compared with a positive pair, which is computationally demanding. In this paper, we propose a different direction and a new loss function for SSL, which is based on the {\em whitening} of the latent-space features. The whitening operation has a ``scattering'' effect on the batch samples,  avoiding degenerate solutions where all the sample representations collapse to a single point. Our solution  does not require asymmetric networks and it is conceptually simple. Moreover, since negatives are not needed, we can extract multiple positive pairs from the same image instance. 
The source code of the method and of all the experiments is
available at: \url{https://github.com/htdt/self-supervised}.
\end{abstract}

\section{Introduction}
\label{Introduction}

One of the current main bottlenecks in deep network training  is the dependence on large annotated training datasets, and this motivates the recent surge of interest in unsupervised methods. Specifically, in self-supervised representation learning (SSL), a network is (pre-)trained without any form of manual annotation, thus providing a means to extract information from unlabeled-data sources  (e.g., text corpora, videos,  images from the Internet, etc.). 
In self-supervision, label-based information is replaced by a prediction problem using some form of {\em context} or using a {\em pretext} task. Pioneering work in this direction was done in Natural Language Processing (NLP), in which the co-occurrence of words in a sentence is used to learn a language model \citep{DBLP:journals/corr/abs-1301-3781,DBLP:conf/nips/MikolovSCCD13,devlin-etal-2019-bert}.
In Computer Vision, typical  contexts or pretext tasks are based on: (1) the temporal consistency in videos 
\citep{DBLP:conf/iccv/WangG15,DBLP:conf/eccv/MisraZH16,DBLP:conf/cvpr/DwibediATSZ19}, (2) the spatial order of patches in still images \citep{DBLP:conf/eccv/NorooziF16,misra2019selfsupervised,CPC2} or (3) simple image transformation techniques \citep{IIC,he2019momentum,wu2018unsupervised}.
The intuitive idea behind most of these methods is to collect pairs of {\em positive} and {\em negative} samples:
 two positive samples should share the same semantics, while negatives should be perceptually different.
A triplet loss \citep{DBLP:conf/nips/Sohn16,DBLP:conf/cvpr/SchroffKP15,Hermans2017InDO,DBLP:conf/iccv/WangG15,DBLP:conf/eccv/MisraZH16} can then be used to learn a metric space representing  the human perceptual similarity. However, most of the recent studies use a  contrastive loss \citep{DBLP:conf/cvpr/HadsellCL06}  or one of its variants \citep{pmlr-v9-gutmann10a,CPC,DIM}, while \citet{DBLP:journals/corr/abs-1907-13625} show the relation between the triplet and the contrastive losses.

It is worth noticing that the success of both kinds of losses is strongly affected by the number and the quality of the negative samples. For instance, in the case of the triplet loss, a common practice is to select {\em hard/semi-hard} negatives \citep{DBLP:conf/cvpr/SchroffKP15,Hermans2017InDO}. On the other hand, \citet{DIM} have shown that the contrastive loss needs a large number of negatives to be competitive. This implies using  batches with a large size, which is computationally demanding, especially with high-resolution images.
In order to alleviate this problem, \citet{wu2018unsupervised} use a {\em memory bank} of negatives, which is composed of feature-vector representations of all the training samples. \citet{he2019momentum} 
conjecture that the use of large and fixed-representation vocabularies is one of the keys to the success of self-supervision in NLP. The solution proposed in MoCo \citet{he2019momentum}  extends \citet{wu2018unsupervised} using a memory-efficient queue of the last visited negatives, together with a {\em momentum encoder} which preserves the intra-queue representation consistency.
\citet{simclr} have performed large-scale experiments confirming that a large number of negatives (and therefore a large batch size) is required for the contrastive loss to be efficient. 

Very recently, \citet{byol} proposed an alternative  direction, in which only positives are used, together with two networks, where the {\em online} network tries to predict the representation of a positive extracted by the {\em target} network. Despite the large success of BYOL \citep{byol},
the reason why the two networks can avoid a collapsed representation (e.g., where all the images are mapped to the same point) is still unclear 
\cite{Byol-collapse-report-1,tian2020understanding,simsiam,richemond2020byol}. According to \citep{Byol-collapse-report-1,tian2020understanding}, one of the important ingredients which is {\em implicitly} used in BYOL to avoid degenerate solutions, is the use of the Batch Norm (BN) \citep{DBLP:conf/icml/IoffeS15} in the projection/prediction heads (see Sec.~\ref{Discussion}). 
In this paper we propose to generalize this finding and we show that, using a full {\em whitening} of the latent space features is {\em sufficient} to avoid collapsed representations, without the need of additional momentum networks  \citep{byol}, siamese networks with stop-gradient operations \cite{simsiam} or the use of specific, batch-based optimizers like LARS \cite{you2017large,Byol-collapse-report-1}.

In more detail, we propose a new SSL loss function, which first {\em scatters} all the sample representations in a spherical distribution\footnote{Here and in the following, with ``spherical distribution'' we mean a distribution with a zero-mean and an identity-matrix covariance.} and then {\em penalizes} the positive pairs which are far from each other. Specifically, given a set of samples $V = \{\mathbf{v}_i\}$, corresponding to the current mini-batch of images $B = \{x_i\}$, we first project the elements of $V$ onto a spherical distribution using a {\em whitening} transform  \citep{siarohin2018whitening}. The whitened representations $\{\mathbf{z}_i\}$, corresponding to $V$, are normalized and then used to compute a Mean Squared Error (MSE) loss which accumulates the error considering only positive pairs $(\mathbf{z}_i,\mathbf{z}_j)$. To avoid a representation collapse, we  do not need to {\em contrast} positives against negatives as in the contrastive loss or in the triplet loss because the optimization process leads to shrinking the distance between positive pairs and, indirectly, scatters the other samples to satisfy the overall spherical-distribution constraint.

In summary, our contributions are the following:

\begin{itemize}
\item
We propose a new SSL loss function, Whitening MSE (W-MSE). W-MSE constrains the batch samples to lie in a spherical distribution and it is an alternative to positive-negative instance contrasting methods.
\item
Our loss does not need a large number of negatives, thus we can include more positives in the current batch. We indeed demonstrate that multiple positive pairs extracted from one image improve the performance.
\item
We empirically show that our W-MSE loss outperforms the commonly adopted contrastive loss and it is competitive with respect to state-of-the-art SSL methods like \cite{byol,simsiam}.
\end{itemize}

\section{Background and Related Work}
\label{RelatedWork}

A typical SSL method is composed of two main components: a {\em pretext task}, which exploits some a-priori knowledge about the domain to automatically extract supervision from data, and a {\em loss function}. 
In this section we briefly review both aspects, and we additionally analyse the recent literature concerning feature whitening.

{\bf Pretext Tasks.}
The temporal consistency in a video provides an intuitive form of self-supervision: temporally-close frames usually contain a similar semantic content \citep{DBLP:conf/iccv/WangG15,CPC}. \citet{DBLP:conf/eccv/MisraZH16} extended this idea using the relative temporal order of 3 frames, while \citet{DBLP:conf/cvpr/DwibediATSZ19} used a {\em temporal cycle consistency} for self-supervision, which is based on comparing two videos sharing the same semantics and computing inter-video frame-to-frame nearest neighbour assignments.

When dealing with still images, the most common pretext task is {\em instance discrimination} (\citet{wu2018unsupervised}): from a training image $x$, a composition of data-augmentation techniques are used to extract two different views of $x$ ($x_i$ and $x_j$). Commonly adopted transformations are:  image cropping, rotation, color jittering, Sobel filtering, etc. The learner, which is usually composed of an encoder and a projection head, is then required to discriminate $(x_i, x_j)$ from other views extracted from other samples \citep{wu2018unsupervised,IIC,he2019momentum,simclr}.
 
Denoising auto-encoders \citep{denoising} add random noise to the input image and try to recover the original image.
\citet{DBLP:conf/eccv/XuCMJT20} enforce  consistency across  different image resolutions.
More sophisticated pretext tasks consist in predicting the spatial order of image patches  \citep{DBLP:conf/eccv/NorooziF16,misra2019selfsupervised} or in reconstructing large masked regions of the image \citep{DBLP:journals/corr/PathakKDDE16}. \citet{DIM,AMDIM} compare the holistic representation of an input image with a patch of the same image. \citet{CPC2} use a similar idea, where the comparison depends on the patch order: the appearance of a given patch should be predicted given the appearance of the patches which lie above it in the image.

We use standard data augmentations   \citep{simclr} to get positive pairs, which is a simple solution   and does not require a pretext-task specific network architecture \citep{DIM,AMDIM,CPC2}.

{\bf Loss functions.} 
Denoising auto-encoders use a {\em reconstruction loss} which compares the generated image with the input image before adding noise. Other generative methods use an {\em adversarial loss} in which a discriminator provides supervisory information to the generator \citep{DBLP:conf/iclr/DonahueKD17,DBLP:conf/nips/DonahueS19}.

Early SSL (deep) discriminative methods used a {\em triplet loss} 
\citep{DBLP:conf/iccv/WangG15,DBLP:conf/eccv/MisraZH16}:
given two {\em positive} images $x_i, x_j$  and a {\em negative} $x_k$ (Sec.~\ref{Introduction}),
together with their corresponding latent-space representations $\mathbf{z}_i,  \mathbf{z}_j, \mathbf{z}_k$, this loss penalizes those cases in which $\mathbf{z}_i$ and $\mathbf{z}_k$ are closer to each other than $\mathbf{z}_i$ and $\mathbf{z}_j$ plus a margin $m$:

\begin{equation}
\label{eq.Triplet}
    L_{Triplet} = - \max (\mathbf{z}_i^T  \mathbf{z}_k - \mathbf{z}_i^T  \mathbf{z}_j + m, 0).
\end{equation}

Most of the recent SSL  discriminative methods are based on some {\em contrastive loss} \citep{DBLP:conf/cvpr/HadsellCL06} variant, in which $\mathbf{z}_i$ and $\mathbf{z}_j$  are contrasted against a set of negative pairs. Following the common formulation proposed by \citet{CPC}, the contrastive loss is given by:

\begin{equation}
\label{eq.InfoNCE}
    L_{Contrastive} = - \log \frac{\exp{(\mathbf{z}_i^T  \mathbf{z}_j / \tau)}}{  \sum_{k=1, k \neq i}^K \exp{(\mathbf{z}_i^T  \mathbf{z}_k / \tau) } },
\end{equation}

\noindent
 where $\tau$ is a {\em temperature} hyperparameter which should be manually set and the sum in the denominator is over a set of $K-1$ negative samples. Usually $K$ is the size of the current batch, i.e., $K = 2 N$, being $N$ the number of  positive pairs. However, as shown by \citet{DIM}, the contrastive loss (\ref{eq.InfoNCE}) requires a large number of negative samples to be competitive. \citet{wu2018unsupervised,he2019momentum} use a set of negatives much larger than the current batch, by pre-computing  representations of old samples. SimCLR \citep{simclr} uses a simpler, but   computationally very demanding, solution based on large batches.

 \citet{DBLP:journals/corr/abs-1907-13625} show that the success of the contrastive loss is likely related to learning a metric space, similarly to what happens with a triplet loss, while
 \citet{hypersphere} investigate   the uniformity and the alignment properties of the $L_2$ normalized contrastive loss. In the same paper, the authors propose two new losses ($\mathcal{L}$\textsubscript{uniform} and $\mathcal{L}$\textsubscript{align}) which explicitly optimize these characteristics.
 
In BYOL (\citet{byol}), given a pair of positives $(x_i, x_j)$, $x_i$ is fed to  the  ``online'' network, which should predict the output of the ``target'' network, where the latter receives $x_j$ as input and its  parameters  are a running average of the former. Concurrently with our work, \citet{simsiam} have simplified this scheme, introducing SimSiam, where both samples $x_i$ and $ x_j$ are encoded using the same network (i.e., a Siamese architecture with a shared encoder). Both BYOL and SimSiam include an additional asymmetric prediction head to compare the latent representations of $x_i$ and $ x_j$, a stop-gradient operation with respect to the element of the positive pair not used by the prediction head,
and an MSE loss of the $L_2$-normalized latent representations. In these methods, both the projection  and  the prediction head include BN. \citet{simsiam}  show that these BN layers are a crucial component of SimSiam, and their removal results in a dramatic performance degradation. \citet{Byol-collapse-report-1} and \citet{tian2020understanding} have empirically confirmed the importance of BN in BYOL, and they show that BN allows BYOL to avoid collapsing the  representation of all images to a constant value, which would make the MSE computation equal to zero. Our work can be seen as a generalization of this finding {\em with a much simpler network architecture} and without the need to rely on asymmetric learning protocols
(see Sec.~\ref{Discussion}). Moreover, our loss formulation is simpler also because it does not require a proper setting of hyperparameters such as  $\tau$  in Eq.~\ref{eq.InfoNCE} or $m$ in Eq.~\ref{eq.Triplet}.

Finally, another recent line of work is based on clustering approaches \citep{NIPS2016_65fc52ed,Caron_2018_ECCV,DBLP:journals/corr/abs-1903-12355}.
For instance, SwAV \cite{caron2020unsupervised} computes a cross-entropy loss over the image-to-cluster prototype assignments, and it is one of the state-of-the-art SSL methods. One of the ingredients of SwAV is the use of multiple positives, and in Sec.~\ref{Discussion} we show that our proposal can exploit multiple crops in a more efficient way.

{\bf Feature Whitening.}
We adopt the efficient and stable Cholesky decomposition \citep{Cholesky} based 
{\em whitening} transform proposed by \citet{siarohin2018whitening} to project our latent-space vectors into a spherical distribution (see Sec.~\ref{Method-Whitening}). Note that \citet{DBN,siarohin2018whitening} use whitening transforms in the intermediate layers of the network for a completely different task: extending BN to a multivariate batch normalization.

\begin{figure}[ht]
\centering
\includegraphics[width=\linewidth]{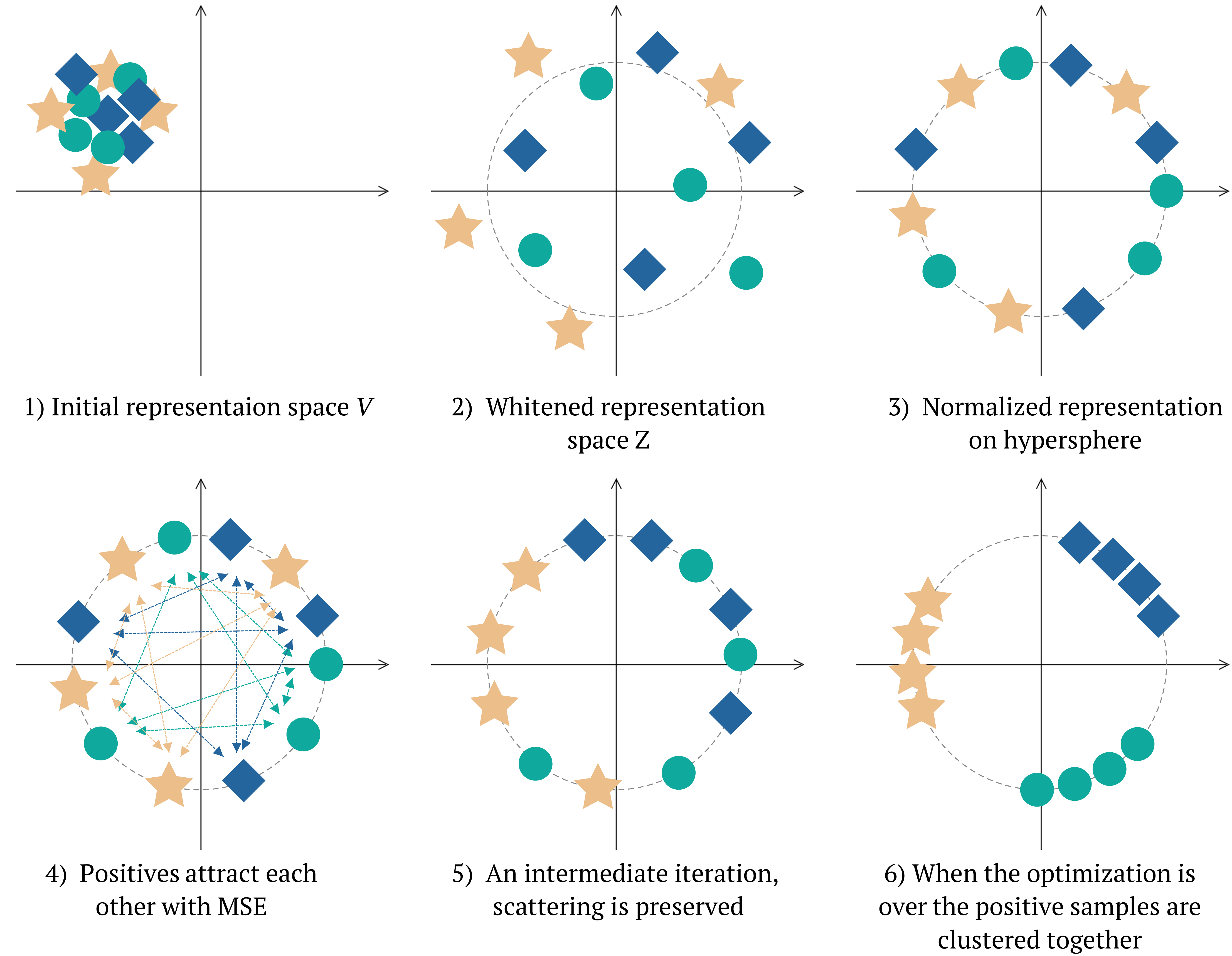}
\caption{A schematic representation of the W-MSE based optimization process. Positive pairs are indicated with the same shapes and colors.
(1) A representation of the batch features in  $V$ when training starts.
(2, 3) The distribution of the elements after whitening and the $L_2$ normalization.
(4) The MSE computed over the normalized $\mathbf{z}$ features encourages the network to move the positive pair representations closer to each other.
(5) The subsequent iterations move closer and closer the positive pairs, while the relative layout of the other samples is forced to lie in a spherical distribution.
}
\label{fig.scheme}
\end{figure}

\section{The Whitening  MSE  Loss}
\label{Method-Whitening}

Given an image $x$, we extract an embedding  $\mathbf{z} = f(x; \theta)$ using an encoder network $f(\cdot; \theta)$ parametrized with $\theta$ (more details below). We require that: (1) the image embeddings are drawn from a non-degenerate distribution (the latter being a distribution where, e.g., all the representations collapse to a single point), and (2) positive image pairs $(x_i, x_j)$, which share a similar semantics, should be clustered close to each  other. We formulate this problem as follows:

\begin{equation}
\label{eq:problem}
min_{\theta}
\mathop{\mathbb{E}}
[dist(\mathbf{z}_i, \mathbf{z}_j)],
\end{equation}
\begin{equation}
\label{eq:constraints}
s.t.~cov(\mathbf{z}_i, \mathbf{z}_i) = cov(\mathbf{z}_j,  \mathbf{z}_j) = I,
\end{equation}

where $dist(\cdot)$ is a distance between vectors, $I$ is the identity matrix and $(\mathbf{z}_i, \mathbf{z}_j)$ corresponds to a positive pair of images $(x_i, x_j)$. With Eq.~\ref{eq:constraints}, we constrain the distribution of the $\mathbf{z}$ values to be non-degenerate, hence avoiding that all the probability mass is concentrated in a single point. Moreover, Eq.~\ref{eq:constraints} makes all the components of $\mathbf{z}$ to be linearly independent from each other, which encourages the different dimensions of $\mathbf{z}$ to represent different semantic content. 
We define the distance with the cosine similarity, implemented with MSE between normalized vectors:
\begin{equation}
\label{eq:dist}
dist(\mathbf{z}_i, \mathbf{z}_j) = \norm{ \frac{\mathbf{z}_i}{\norm{\mathbf{z}_i}_2} - \frac{\mathbf{z}_j}{\norm{\mathbf{z}_j}_2} }^2_2
\end{equation}
\begin{equation*}
= 2 - 2 \frac{\langle \mathbf{z}_i, \mathbf{z}_j \rangle}{\norm{\mathbf{z}_i}_2 \cdot \norm{\mathbf{z}_j}_2}
\end{equation*}

In the Appendix we also include other experiments in which the cosine similarity is replaced by the  Euclidean distance. We provide below the details on how positive image samples are collected, how they are encoded and how the above optimization is implemented.

First, similarly to \citet{simclr}, we obtain positive samples sharing the same semantics from a single image $x$ and using standard image transformation techniques. Specifically, we use a composition of image cropping, grayscaling and color jittering transformations $T(\cdot; \mathbf{p})$. The parameters ($\mathbf{p}$) are selected uniformly at random and independently for each  {\em positive} sample extracted from the same image: $x_{i} = T(x; \mathbf{p}_{i})$. We concisely indicate with $pos(i,j)$ the fact that $x_i$ and $x_j$ ($x_i, x_j \in B$, $B$ the current batch)  have been extracted from the same image.

The number of positive samples per image $d$ may vary, trading off diversity in the batch and the amount of the training signal. Favoring more negatives, most of the methods use only one positive pair ($d = 2$). However, \citet{IIC} have demonstrated improved performance with 5 samples, while \citet{caron2020unsupervised} use 8 samples. In our MSE-based loss (see below), we use all the possible $d (d - 1) / 2$ combinations of positive samples. We include experiments for $d = 2$ (1 positive pair) and $d = 4$ (6 positive pairs). Note that our implementation includes {\em batch slicing}  (described below), where the choice of $d$
is related with the sub-batch size, and  larger  $d$ values can produce instability issues.

\begin{figure*}[ht]
\centering
\includegraphics[width=.75\linewidth]{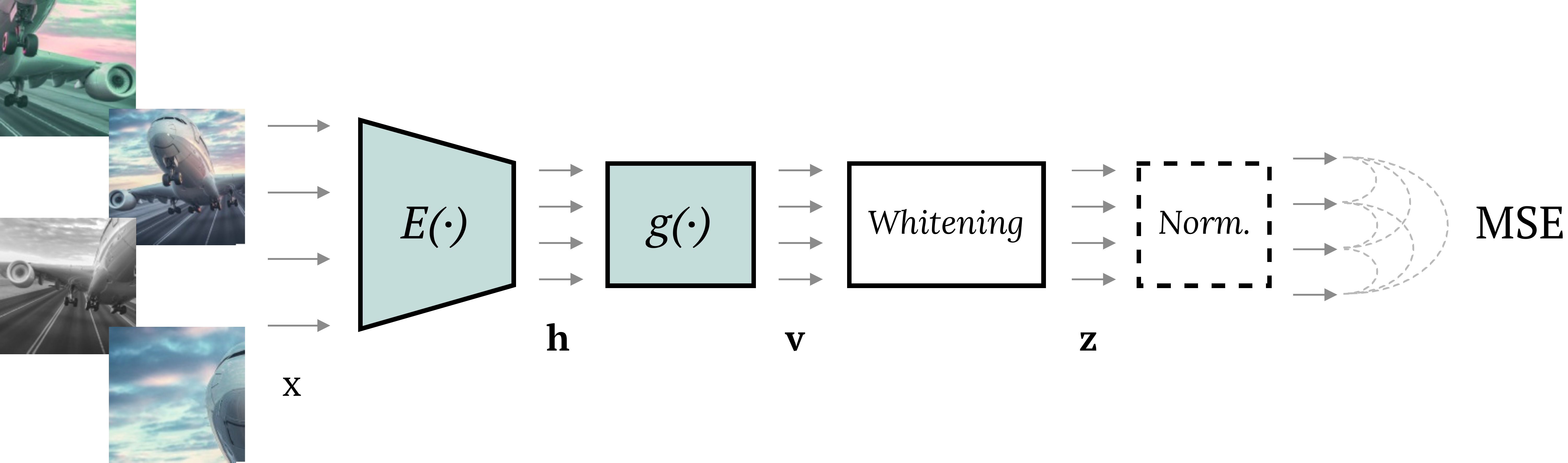}
\caption{A scheme of our training procedure. First, $d$ ($d =4$ in this case) positive samples are generated using augmentations. These images are transformed into vectors with the encoder $E(\cdot)$. Next, they are projected onto a lower dimensional space with a projection head $g(\cdot)$. Then, Whitening projects these vectors onto a spherical distribution, optionally followed by an  $L_2$ normalization. Finally, the dashed curves  show all the  $d (d - 1) / 2$ comparisons used in our W-MSE loss.}
\label{fig.encoder}
\end{figure*}

For representation learning, we use a backbone {\em encoder} network $E(\cdot)$. $E(\cdot)$, trained without human supervision, will be used in  Sec.~\ref{Experiments} for evaluation using standard protocols. We use a standard ResNet \citep{DBLP:conf/cvpr/HeZRS16} as the encoder, and $\mathbf{h} = E(x)$ is the  output of the average-pooling layer. This choice has the advantage to be simple and easily reproducible, in contrast to other methods which use encoder architectures specific for a given pretext task (see Sec.~\ref{RelatedWork}). Since $\mathbf{h} \in \mathbb{R}^{512}$ or $\mathbf{h} \in \mathbb{R}^{2048}$ is a high-dimensional vector, following \citet{simclr} we use a nonlinear projection head $g(\cdot)$ to project $\mathbf{h}$ in a lower dimensional space: 
 $\mathbf{v} = g(\mathbf{h})$, where $g(\cdot)$ is implemented with a MLP with one hidden layer and a BN layer. The whole network $f(\cdot)$ is given by the composition of $g(\cdot)$ with $E(\cdot)$ (see Fig.~\ref{fig.encoder}). Note that we do not use prediction heads like in \cite{byol,simsiam}. 

 Given $N$ original images and a batch of samples $B = \{ x_1, ... x_K \}$, where $K = N d$, let $V = \{ \mathbf{v}_1, ... \mathbf{v}_K \}$, be the corresponding batch of features obtained as described above. In the proposed W-MSE loss, we compute the MSE  over all $N d (d - 1) / 2$ positive pairs, where constraint~\ref{eq:constraints} is satisfied using the reparameterization of the $\mathbf{v}$ variables with the whitened variables $\mathbf{z}$:

 \begin{equation}
 \label{eq.WMSE}
     L_{W\--MSE} (V) = \frac{2}{N d (d-1)}  \sum dist(\mathbf{z}_i, \mathbf{z}_j),
 \end{equation}

\noindent

where the sum is over $(\mathbf{v}_i,\mathbf{v}_j) \in V$, $pos(i,j) = true$, $\mathbf{z} = Whitening(\mathbf{v})$, and:

\begin{equation}
\label{eq.whitening}
Whitening(\mathbf{v}) = W_V (\mathbf{v} - \boldsymbol{\mu}_V). 
\end{equation}

\noindent
In Eq.~\ref{eq.whitening},  $\boldsymbol{\mu}_V$ is the mean of the elements in $V$:
$\boldsymbol{\mu}_V = \frac{1}{K} \sum_k \mathbf{v}_k$, 
while the matrix $W_V$ is such that: $W_V^\top W_V = \Sigma_V^{-1}$, being $\Sigma_V$ the covariance matrix of $V$:

\begin{equation}
\label{eq.sigma-V}
\Sigma_V = \frac{1}{K-1} \sum_k (\mathbf{v}_k - \boldsymbol{\mu}_V) (\mathbf{v}_k - \boldsymbol{\mu}_V)^T. 
\end{equation}

For more details on how $W_V$ is computed, we refer to the Appendix. Equation~\ref{eq.whitening} performs the full whitening of each $\mathbf{v}_i \in V$ and the resulting set of vectors $Z = \{ \mathbf{z}_1, ..., \mathbf{z}_K \}$ lies in a  zero-centered distribution  with a covariance matrix equal to the identity matrix (Fig.~\ref{fig.scheme}).

The intuition behind the proposed loss is that Eq.~\ref{eq.WMSE} penalizes positives which are far apart from each other, thus leading $g(E(\cdot))$ to shrink the inter-positive distances. On the other hand, since $Z$ must lie in a spherical distribution, the other samples should be ``moved'' and rearranged in order to satisfy constraint~\ref{eq:constraints} (see Fig.~\ref{fig.scheme}).

\begin{figure}[ht]
\centering
\includegraphics[width=1\linewidth]{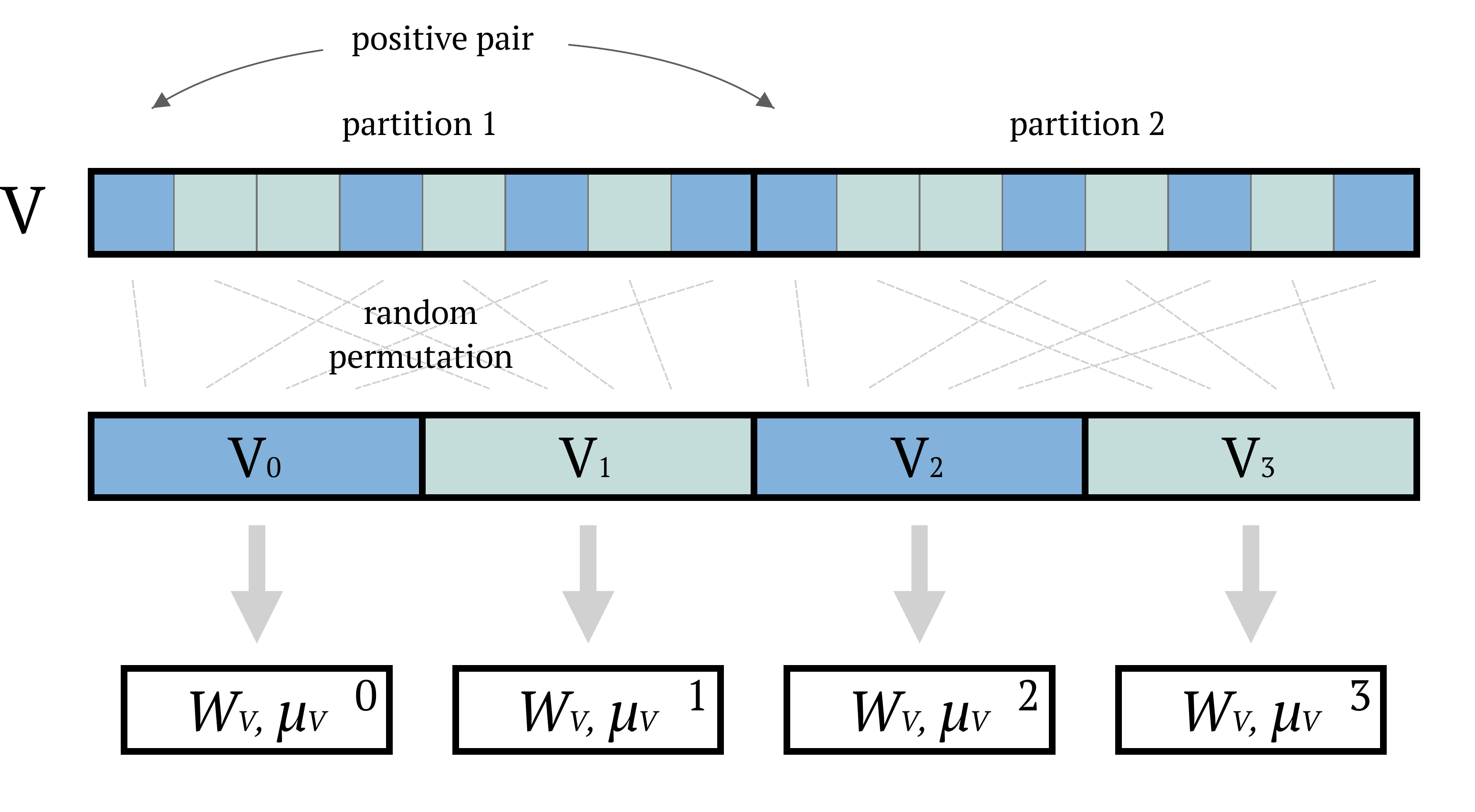}
\caption{Batch slicing. $V$ is first partitioned in $d$ parts ($d=2$ in this example). We randomly permute the first part and we apply the same permutation to the other $d-1$ parts. Then, we further split all the partitions and we create sub-batches ($V_i$). Each $V_i$ is independently used to compute the sub-batch specific  whitening matrix $W_V^i$ and centroid $\boldsymbol{\mu}_V^i$.}
\label{fig.slicing}
\end{figure}

{\bf Batch Slicing.} The estimation of the MSE in Eq.~\ref{eq.WMSE} depends on the whitening matrix $W_V$, which may have a high variance over consecutive-iteration  batches $V_t, V_{t+1}, ...$. For this reason, inspired by the resampling methods~\citep{efron1982jackknife}, given a batch $V$, we slice $V$ in different non-overlapping sub-batches and we compute a whitening matrix independently for each sub-batch. 
In more detail, we first partition the batch in $d$ parts, being $d$ the number of positives extracted from one image. In this way,  each partition contains elements extracted from different original images (i.e., no pair of positives is included in a single partition, see Fig.~\ref{fig.slicing}). 
Then, we randomly permute the elements of the each partition, using the same  permutation for all the partitions. Next, each partition is further split in sub-batches, using the heuristic that the size of each sub-batch ($V_i$) should be equal to the size of embedding ($|\mathbf{v}|$) times 2 (this prevents instability issues when computing the covariance matrices).
For each $V_i$, we use only its elements to compute a corresponding whitening matrix $W_V^i$, which is used to whiten the elements of $V_i$ only (Fig.~\ref{fig.slicing}). In the loss computation (Eq.~\ref{eq.WMSE}), all the elements of all the sub-batches are used, thus implicitly alleviating the differences among the different whitening matrices.
Finally, it is possible to repeat the whole operation several times and to average the result to get a more robust estimate of Eq.~\ref{eq.WMSE}.
Note that, despite  using smaller batches may increase the
instability in computing the whitening matrix,  this is compensated
by having a different $W_V^i$ for each $V_i$, and by the
possibility to iterate the sampling process with different permutations
for the same $V$.

\subsection{Discussion}
\label{Discussion}

In a common {\em instance-discrimination} task (Sec.~\ref{RelatedWork}), e.g., solved using Eq.~\ref{eq.InfoNCE}, the similarity of a positive pair ($\mathbf{z}_i^T  \mathbf{z}_j$) is contrasted with the similarity computed with respect to all the other samples ($\mathbf{z}_k$) in the batch ($\mathbf{z}_i^T  \mathbf{z}_k$, $1 \leq k \leq K, k \neq i$). 
 However, $\mathbf{z}_k$ and $\mathbf{z}_i$, extracted from different image instances, can occasionally share the same semantics (e.g., $x_i$ and $x_k$ are two different image {\em instances} of the unknown ``cat'' {\em class}). Conversely, the proposed W-MSE loss 
does not force all the  instance samples to lie far from each other, but it only imposes a soft constraint (Eq.~\ref{eq:constraints}), which avoids degenerate distributions.

Note that previous work \citep{he2019momentum,CPC2,simclr} highlighted that BN may be harmful for learning semantically meaningful representations because the network can ``cheat'' and exploit the batch statistics in order to find a trivial solution to Eq.~\ref{eq.InfoNCE}. However, our whitening transform (Eq.~\ref{eq.whitening}) is applied only to  the very last layer of the network $f(\cdot)$ (see Fig.~\ref{fig.encoder}) and it is not used in the intermediate layers, which is instead  the case of BN. Hence, our $f(\cdot)$ cannot learn to exploit subtle inter-sample dependencies introduced by batch-statistics because of the lack of other learnable layers on top of the $\mathbf{z}$ features.

 Similarly to Eq.~\ref{eq.WMSE}, in BYOL \citep{byol} an MSE loss is used to compare the latent representations of two positives computed by slightly different networks without contrasting positives with negatives 
 (Sec.~\ref{RelatedWork}). However, 
 the MSE loss alone could be trivially minimized by a collapsed distribution, where both networks output sample-independent representations. The reason why this does not happen is still an open problem and 
 this has opened a debate in the community
 \cite{Byol-collapse-report-1,tian2020understanding,simsiam,richemond2020byol}. Specifically, it seems that the BN layers, included in the projection/prediction heads of BYOL are {\em one of the} ingredients which help to avoid degenerate solutions \citep{Byol-collapse-report-1,tian2020understanding} (see Sec.~\ref{Introduction} and \ref{RelatedWork}).
  Our W-MSE can be seen as a generalization of this implicit property of BYOL, in which the $\mathbf{z}$ values of the current batch are full-whitened, so preventing possible collapsing effects of the MSE loss. Importantly, we reach this result without 
  the need of  specific asymmetric architectures (e.g., momentum networks \citep{byol}, prediction heads, etc.) or sophisticated 
  training protocols (e.g., stop-gradient, slow-convergence of the prediction head \citep{simsiam}, etc.).

Finally, note that using BN alone without whitening (Eq.~\ref{eq.whitening}), and without the aforementioned asymmetric learning, is {\em not sufficient}. Indeed, if we  minimize the MSE after feature standardization (i.e., BN), the network can easily find a solution where, e.g.,  all the dimensions of the embedding represent the same feature or are correlated to each other. 
For instance, in  preliminary experiments with CIFAR-10,
we replaced whitening with BN, and the network converged to a 0 loss value after 50 epochs. Nevertheless, the linear classification accuracy was very poor: $68.15\%$. 

Last but not least, the use of multiple positives extracted from the same image (i.e., $d > 2$) has been  recently proposed in SwAV \citep{caron2020unsupervised}.
However, different from our proposal, in SwAV, the multi-crop strategy is based on multiple-resolution crops, and, most importantly, in our case, we can compute $d (d-1)$ inter-positive differences in Eq.~\ref{eq.WMSE}  with only $d$ forward passes, while, in SwAV, the number of comparisons grows linearly with $d$. 

\section{Experiments}
\label{Experiments}

In our experiments we use the following {\bf datasets}. 

\begin{itemize}
\item
CIFAR-10 and CIFAR-100 \citep{CIFAR}, two small-scale  datasets composed of $32 \times 32$ images with 10 and 100 classes, respectively.

\item
ImageNet \citep{imagenet_cvpr09}, the well-known  large-scale dataset with about 1.3M training images and 50K test images, spanning over 1000 classes.

\item
Tiny ImageNet \citep{tinyin}, a reduced version of ImageNet, composed of 200 classes with images scaled down to $64 \times 64$. The total number of  images is: 100K (training) and 10K (testing). 

\item
ImageNet-100  \citep{DBLP:conf/eccv/TianKI20}, a random 100-class subset of ImageNet.

\item
STL-10 \citep{DBLP:journals/jmlr/CoatesNL11}, also  derived from ImageNet, with $96 \times 96$ resolution images and more than 100K training samples.

\end{itemize}

{\bf Setting.}
The goal of our experiments is to compare W-MSE with state-of-the-art SSL losses, isolating the effects of other settings, such as the architectural choices. For this reason, in the small and medium size dataset experiments of Tab.~\ref{table.sota}, we use the same encoder $E(\cdot)$, ResNet-18, for all the compared methods and, similarly, we use  ResNet-50 for the ImageNet-based experiments in Tab.~\ref{table.imagenet}. When we do not report previously published results, we independently select the best hyperparameter values for each method and each dataset. In each method, the latent-space   features are $L_2$ normalized, unless otherwise specified. 
In Tab.~\ref{table.sota},
{\em SimCLR (our repro.)} refers to our implementation of the contrastive loss (Eq.~\ref{eq.InfoNCE}) following the details in \citep{simclr}, with temperature $\tau = 0.5$. In the same table, {\em BYOL (our repro.)} is our reproduction of \citep{byol}. For this method we use the exponential moving average with cosine increasing, starting from $0.99$. {\em W-MSE 2} and {\em W-MSE 4} correspond to our method  with $d = 2$ and $d = 4$ positives extracted per image, respectively. For CIFAR-10 and CIFAR-100, the slicing sub-batch size is $128$, for Tiny ImageNet and STL-10, it is $256$. In the Tiny ImageNet and STL-10 experiments with W-MSE 2,   we use $4$ iterations of batch slicing, while in all the other experiments we use only $1$ iteration.

{\bf Implementation details.}
For the  small and medium size datasets, we use the Adam optimizer \citep{kingma2014adam}. For all the compared methods (including ours), we use the same number of epochs and the same learning rate schedule. Specifically, for CIFAR-10 and CIFAR-100, we use 1,000 epochs with learning rate $3\times10^{-3}$; for Tiny ImageNet, 1,000 epochs with learning rate $2\times10^{-3}$; for STL-10, 2,000 epochs with learning rate $2\times10^{-3}$. We use learning rate warm-up for the first 500 iterations of the optimizer, and a $0.2$ learning rate drop $50$ and $25$ epochs before the end. We use a mini-batch size of $K = 1024$ samples. The dimension of the hidden layer of the projection head $g(\cdot)$ is 1024. The weight decay is $10^{-6}$. Finally, we use an embedding size of $64$ for CIFAR-10 and CIFAR-100, and an embedding of size of $128$ for STL-10 and Tiny ImageNet. For ImageNet-100 we use a configuration similar to the Tiny ImageNet experiments, and 240 epochs of training.
Finally, in the ImageNet experiments (Tab.~\ref{table.imagenet}),  we 
use the   implementation and the hyperparameter configuration
 of \citep{simclr_v2} (same  number of layers in the projection head, etc.)
 based on their open-source implementation\footnote{\url{https://github.com/google-research/simclr}}, the only difference being the learning rate  and the loss function (respectively, 0.075 and the contrastive loss in \citep{simclr_v2} vs. 0.1 and Eq.~\ref{eq.WMSE} in W-MSE 4).


{\bf Image Transformation Details.}
We extract crops with a random size from $0.2$ to $1.0$ of the original area and a random aspect ratio from $3/4$ to $4/3$ of the original aspect ratio, which is a commonly used data-augmentation technique \citep{simclr}. We also apply horizontal mirroring with probability $0.5$. Finally, we apply color jittering with configuration ($0.4, 0.4, 0.4, 0.1$) with probability $0.8$ and grayscaling with probability $0.1$. For ImageNet and ImageNet-100, we follow the details in \citep{simclr}: crop size from $0.08$ to $1.0$, stronger jittering ($0.8, 0.8, 0.8, 0.2$), grayscaling probability $0.2$, and Gaussian blurring with $0.5$ probability.
In all the experiments, at testing time we use only one
crop (standard  protocol).

{\bf Evaluation Protocol.} The most common evaluation protocol for unsupervised feature learning is based on {\em freezing} the network encoder ($E(\cdot)$, in our case) after unsupervised pre-training, and then train a supervised {\em linear classifier} on top of it. Specifically, the linear classifier is a fully-connected layer followed by softmax, which is plugged on top of $E(\cdot)$ after removing the projection head $g(\cdot)$. In all the experiments, we train the linear classifier for 500 epochs using the Adam optimizer and the labeled training set of each specific dataset, without data augmentation. The learning rate is exponentially decayed from $10^{-2}$ to $10^{-6}$. The weight decay is $5 \times 10^{-6}$.
In our experiments, we also include the accuracy of a k-nearest neighbors classifier (k-nn, $k=5$). The advantage of using this classifier  is that it does not require additional parameters and training, and it is deterministic.

\subsection{Comparison with the state of the art}
\label{exp.sota}

\begin{table*}[t]
\begin{center}
\caption{Classification accuracy (top 1) of a linear classifier and a 5-nearest neighbors classifier for different loss functions and datasets with a ResNet-18 encoder.}
\label{table.sota}
{\renewcommand{\arraystretch}{1.25}
\begin{tabular}{l|r r|r r|r r|r r}
\toprule
Method & \multicolumn{2}{c |}{CIFAR-10} & \multicolumn{2}{c |}{CIFAR-100} & \multicolumn{2}{c |}{STL-10} & \multicolumn{2}{c}{Tiny ImageNet} \\
 & linear & 5-nn & linear & 5-nn & linear & 5-nn & linear & 5-nn \\
\midrule
SimCLR \citep{simclr} (our repro.) & 91.80 & 88.42 & 66.83 & 56.56 & 90.51 & 85.68 & 48.84 & 32.86 \\
BYOL \citep{byol} (our repro.)  & 91.73 & 89.45 & 66.60 & {\bf 56.82} & {\bf 91.99} & {\bf 88.64} & {\bf 51.00} & {\bf 36.24} \\
W-MSE 2   (ours)      & 91.55 & 89.69 & 66.10 & 56.69 & 90.36 & 87.10 & 48.20 & 34.16 \\
W-MSE 4   (ours)      & {\bf 91.99} & {\bf 89.87} & {\bf 67.64} & 56.45 & 91.75 & 88.59 & 49.22 & 35.44 \\
\bottomrule
\end{tabular}}
\end{center}
\end{table*}

\begin{table}[t]
\begin{center}
\caption{Classification accuracy on ImageNet-100. Top 1 and 5 correspond to the accuracy of a linear classifier. W-MSE (2 and 4) are based on a {\em ResNet-18} encoder. \textsuperscript{\dag}~indicates that the results are based on a {\em ResNet-50} encoder and the values are reported from \citep{hypersphere}.}
\label{table.in100}
{\renewcommand{\arraystretch}{1.25}
\begin{tabular}{l|c c c}
\toprule
Method & top 1 & top 5 & 5-nn \\
\midrule
MoCo \citep{he2019momentum} 
\textsuperscript{\dag} & 72.80 & 91.64 & - \\
$\mathcal{L}$ \textsubscript{align} and $\mathcal{L}$\textsubscript{uniform} \\
\citep{hypersphere} 
\textsuperscript{\dag} & 74.60 & 92.74 & - \\
W-MSE 2 (ours) & 76.00 & 93.14 & 67.04 \\
W-MSE 4 (ours) & {\bf 79.02} & {\bf 94.46} & {\bf 71.32} \\
\bottomrule
\end{tabular}}
\end{center}
\end{table}

\begin{table}[t]
\begin{center}
\caption{Classification accuracy (top 1) of a linear classifier on ImageNet with a ResNet-50 encoder. All results but ours are reported from \citep{simsiam}. \textsuperscript{\dag} The reproduction of SwAV in \citep{simsiam} does not include a multi-crop strategy.}
\label{table.imagenet}
{\renewcommand{\arraystretch}{1.25}
\begin{tabular}{l|c|c}
\toprule
Method & 100 epochs & 400 epochs \\
\midrule
SimCLR \citep{simclr}  & 66.5  & 69.8 \\
MoCo v2 \citep{DBLP:journals/corr/abs-2003-04297}     &  67.4 & 71.0 \\
BYOL \citep{byol}    &  66.5 & {\bf 73.2} \\
SwAV  \citep{caron2020unsupervised} \textsuperscript{\dag} &  66.5 & 70.7 \\
SimSiam \citep{simsiam}   & 68.1 & 70.8 \\
W-MSE 4 (ours) & {\bf 69.43}  & 72.56 \\
\bottomrule
\end{tabular}}
\end{center}
\end{table}

Tab.~\ref{table.sota} shows the results of the experiments on small and medium size datasets. For W-MSE, 4 samples are generally better than 2. The contrastive loss performs the worst in most cases. The W-MSE 4 accuracy is the best on CIFAR-10 and CIFAR-100, while BYOL leads on STL-10 and Tiny ImageNet, although the gap between the two methods is marginal. In the Appendix, we plot the linear classification accuracy during training  for the STL-10 dataset. The plot shows that W-MSE 4 and BYOL have a similar performance during most of the training. However, in the first 120 epochs, BYOL significantly underperforms W-MSE 4 (e.g., the accuracy after 20 epochs: W-MSE 4, $79.98\%$; BYOL, $73.24\%$), indicating that BYOL requires a ``warmup'' period. 
On the other hand, W-MSE performs well from the beginning. This property is useful in those domains which require a rapid adaptation of the encoder, e.g., due to the change of the data distribution in continual learning or in reinforcement learning.

Tab.~\ref{table.in100} shows the results on a larger dataset (ImageNet-100). In that table, MoCo is the contrastive-loss based method proposed in
\citep{he2019momentum}, and $\mathcal{L}$\textsubscript{align} and $\mathcal{L}$\textsubscript{uniform} are the two losses proposed in \citep{hypersphere}
(Sec.~\ref{Introduction}-\ref{RelatedWork}). Note that, while W-MSE (2 and 4) in Tab.~\ref{table.in100} refer to our method with a {\em ResNet-18} encoder, the other results are reported from \citep{hypersphere}, where a much larger-capacity network (i.e., a {\em ResNet-50}) is used as the encoder. {\em Despite this large difference in the encoder capacity}, both versions of W-MSE significantly outperform the other two compared methods in this dataset.
Tab.~\ref{table.in100} also shows that W-MSE 2, the version of our method without multi-cropping, is highly competitive, being its classification accuracy significantly higher than state-of-the-art methods.

Finally, in Tab.~\ref{table.imagenet} we show the ImageNet results  using 100 and 400 training epochs, and we compare 
W-MSE 4 with the results of other state-of-the-art approaches as reported in \citep{simsiam}. Despite some configuration  details are different (e.g., the depth of the projection head, etc.), in all cases the encoder is a ResNet-50. However, SwAV refers to the reproduction of \cite{caron2020unsupervised} used in \cite{simsiam}, where no multi-crop strategy is adopted (hence, $d=2$), and a multi-crop version of SwAV may likely obtain significantly larger values.
Tab.~\ref{table.imagenet} shows that W-MSE 4 
is the  state of the art with 100 epochs
and it is very close to the 400-epochs state of the art.
These results confirm that our method is
highly competitive, considering that we  have not intensively tuned our
hyperparameters  and that our network is much
simpler than other approaches.

\subsection{Contrastive loss with whitening}
\label{contrastive_white}

\begin{table}[th!]
\begin{center}
\caption{CIFAR-10: accuracy of the  contrastive loss with whitened features, trained for 200 epochs.}
\label{table.contrastive_white}
{\renewcommand{\arraystretch}{1.25}
\begin{tabular}{c c c c}
\toprule
Whitened features & $L_2$ normalized & linear & 5-nn \\
\midrule
\xmark & \cmark & 89.66 & 86.55 \\
\cmark & \cmark & \multicolumn{2}{c}{diverged} \\
\xmark & \xmark & 79.48 & 76.60 \\
\cmark & \xmark & 77.39 & 74.14 \\
\bottomrule
\end{tabular}}
\end{center}
\end{table}

In this section,  we analyse the effect of the whitening transform in combination with  the  contrastive loss. 
Specifically, we use the  contrastive loss (Eq.~\ref{eq.InfoNCE}) on whitened features $\mathbf{z} = Whitening(\mathbf{v})$ (Eq.~\ref{eq.whitening}).
Tab.~\ref{table.contrastive_white} shows the results  on CIFAR-10. The first row refers to the standard contrastive loss without whitening. Note that the difference with respect to Tab.\ref{table.sota} is due to the use of only 200 training epochs. The second row refers to Eq.~\ref{eq.InfoNCE}, where the features ($\mathbf{z}$) are computed using Eq.~\ref{eq.whitening} and then $L_2$ normalized, while in the last two rows, $\mathbf{z}$ is not normalized.
If the features are whitened and then normalized, we observed an unstable training, with divergence after a few epochs. The unnormalized version with whitening converged, but its accuracy is worse than the standard contrastive loss (both normalized and unnormalized). 

These experiments show that the whitening transform alone does not improve the SSL performance, and, 
 used jointly with negative contrasting,  it may be harmful.
Conversely, we use whitening in our W-MSE
to avoid a collapsed representation (Eq.~\ref{eq:constraints})
when only positives are used (Eq.~\ref{eq:problem}).

\section{Conclusion}
\label{Conclusion}

In this paper we  proposed a new SSL  loss, W-MSE, which is alternative to common loss functions used in the field. Differently from the triplet loss and the contrastive loss, both of which are based on comparing a pair of positive instances against other samples, W-MSE computes only the inter-positive distances, while using a whitening transform to avoid degenerate solutions.
Our proposal is similar to recent SSL methods like BYOL \citep{byol} and 
SimSiam \citep{simsiam}, which use only positives to reduce the dependence on large batch sizes \citep{simclr}. However, differently from BYOL and SimSiam,
which adopt asymmetric network architectures and traing protocols, our solution to avoid collapsed representations is much simpler, while achieving 
a classification accuracy which is, in most cases, comparable or superior to  the state-of-the-art methods. 
Despite asymmetry in learning and whitening are
alternative solutions,
a combination is possible. This may be a direction of future investigation.

 \section*{Acknowledgements}
 This work was supported by EU H2020 project AI4Media No.  951911  and by the  EUREGIO project OLIVER.

\bibliography{_main}

\begin{thebibliography}{52}
\providecommand{\natexlab}[1]{#1}
\providecommand{\url}[1]{\texttt{#1}}
\expandafter\ifx\csname urlstyle\endcsname\relax
  \providecommand{\doi}[1]{doi: #1}\else
  \providecommand{\doi}{doi: \begingroup \urlstyle{rm}\Url}\fi

\bibitem[Bachman et~al.(2019)Bachman, Hjelm, and Buchwalter]{AMDIM}
Bachman, P., Hjelm, R.~D., and Buchwalter, W.
\newblock Learning representations by maximizing mutual information across
  views.
\newblock In \emph{NeurIPS}, 2019.

\bibitem[Bautista et~al.(2016)Bautista, Sanakoyeu, Tikhoncheva, and
  Ommer]{NIPS2016_65fc52ed}
Bautista, M.~A., Sanakoyeu, A., Tikhoncheva, E., and Ommer, B.
\newblock {CliqueCNN:} deep unsupervised exemplar learning.
\newblock In \emph{NeurIPS}, 2016.

\bibitem[Caron et~al.(2018)Caron, Bojanowski, Joulin, and
  Douze]{Caron_2018_ECCV}
Caron, M., Bojanowski, P., Joulin, A., and Douze, M.
\newblock Deep clustering for unsupervised learning of visual features.
\newblock In \emph{ECCV}, 2018.

\bibitem[Caron et~al.(2020)Caron, Misra, Mairal, Goyal, Bojanowski, and
  Joulin]{caron2020unsupervised}
Caron, M., Misra, I., Mairal, J., Goyal, P., Bojanowski, P., and Joulin, A.
\newblock Unsupervised learning of visual features by contrasting cluster
  assignments.
\newblock In \emph{NeurIPS}, 2020.

\bibitem[Chen et~al.(2020{\natexlab{a}})Chen, Kornblith, Norouzi, and
  Hinton]{simclr}
Chen, T., Kornblith, S., Norouzi, M., and Hinton, G.
\newblock A simple framework for contrastive learning of visual
  representations.
\newblock \emph{arXiv:2002.05709}, 2020{\natexlab{a}}.

\bibitem[Chen et~al.(2020{\natexlab{b}})Chen, Kornblith, Swersky, Norouzi, and
  Hinton]{simclr_v2}
Chen, T., Kornblith, S., Swersky, K., Norouzi, M., and Hinton, G.
\newblock Big self-supervised models are strong semi-supervised learners.
\newblock \emph{arXiv:2006.10029}, 2020{\natexlab{b}}.

\bibitem[Chen \& He(2020)Chen and He]{simsiam}
Chen, X. and He, K.
\newblock Exploring simple siamese representation learning.
\newblock \emph{arXiv:2011.10566}, 2020.

\bibitem[Chen et~al.(2020{\natexlab{c}})Chen, Fan, Girshick, and
  He]{DBLP:journals/corr/abs-2003-04297}
Chen, X., Fan, H., Girshick, R.~B., and He, K.
\newblock Improved baselines with momentum contrastive learning.
\newblock \emph{arXiv:2003.04297}, 2020{\natexlab{c}}.

\bibitem[Coates et~al.(2011)Coates, Ng, and Lee]{DBLP:journals/jmlr/CoatesNL11}
Coates, A., Ng, A.~Y., and Lee, H.
\newblock An analysis of single-layer networks in unsupervised feature
  learning.
\newblock In \emph{AISTATS}, 2011.

\bibitem[Deng et~al.(2009)Deng, Dong, Socher, Li, Li, and
  Fei-Fei]{imagenet_cvpr09}
Deng, J., Dong, W., Socher, R., Li, L.-J., Li, K., and Fei-Fei, L.
\newblock {ImageNet: A Large-Scale Hierarchical Image Database}.
\newblock In \emph{CVPR09}, 2009.

\bibitem[Dereniowski \& Marek(2004)Dereniowski and Marek]{Cholesky}
Dereniowski, D. and Marek, K.
\newblock Cholesky factorization of matrices in parallel and ranking of graphs.
\newblock In \emph{5th Int. Conference on Parallel Processing and Applied
  Mathematics}, 2004.

\bibitem[Devlin et~al.(2019)Devlin, Chang, Lee, and
  Toutanova]{devlin-etal-2019-bert}
Devlin, J., Chang, M.-W., Lee, K., and Toutanova, K.
\newblock {BERT}: Pre-training of deep bidirectional transformers for language
  understanding.
\newblock In \emph{NAACL,}, 2019.

\bibitem[Donahue \& Simonyan(2019)Donahue and
  Simonyan]{DBLP:conf/nips/DonahueS19}
Donahue, J. and Simonyan, K.
\newblock Large scale adversarial representation learning.
\newblock In \emph{NeurIPS}, 2019.

\bibitem[Donahue et~al.(2017)Donahue, Kr{\"{a}}henb{\"{u}}hl, and
  Darrell]{DBLP:conf/iclr/DonahueKD17}
Donahue, J., Kr{\"{a}}henb{\"{u}}hl, P., and Darrell, T.
\newblock Adversarial feature learning.
\newblock In \emph{ICLR}, 2017.

\bibitem[Dwibedi et~al.(2019)Dwibedi, Aytar, Tompson, Sermanet, and
  Zisserman]{DBLP:conf/cvpr/DwibediATSZ19}
Dwibedi, D., Aytar, Y., Tompson, J., Sermanet, P., and Zisserman, A.
\newblock Temporal cycle-consistency learning.
\newblock In \emph{CVPR}, 2019.

\bibitem[Efron(1982)]{efron1982jackknife}
Efron, B.
\newblock \emph{The jackknife, the bootstrap, and other resampling plans},
  volume~38.
\newblock Siam, 1982.

\bibitem[Fetterman \& Albrecht(2020)Fetterman and
  Albrecht]{Byol-collapse-report-1}
Fetterman, A. and Albrecht, J.
\newblock Understanding self-supervised and contrastive learning with bootstrap
  your own latent {(BYOL)}.
\newblock
  \emph{https://untitled-ai.github.io/understanding-self-supervised-contrastive-learning.html},
  2020.

\bibitem[Grill et~al.(2020)Grill, Strub, Altché, Tallec, Richemond,
  Buchatskaya, Doersch, Pires, Guo, Azar, Piot, Kavukcuoglu, Munos, and
  Valko]{byol}
Grill, J.-B., Strub, F., Altché, F., Tallec, C., Richemond, P.~H.,
  Buchatskaya, E., Doersch, C., Pires, B.~A., Guo, Z.~D., Azar, M.~G., Piot,
  B., Kavukcuoglu, K., Munos, R., and Valko, M.
\newblock Bootstrap your own latent: A new approach to self-supervised
  learning.
\newblock \emph{arXiv:2006.07733}, 2020.

\bibitem[Gutmann \& Hyvärinen(2010)Gutmann and Hyvärinen]{pmlr-v9-gutmann10a}
Gutmann, M. and Hyvärinen, A.
\newblock {Noise-contrastive estimation: A new estimation principle for
  unnormalized statistical models}.
\newblock In \emph{Proceedings of the Thirteenth International Conference on
  Artificial Intelligence and Statistics}, 2010.

\bibitem[Hadsell et~al.(2006)Hadsell, Chopra, and
  LeCun]{DBLP:conf/cvpr/HadsellCL06}
Hadsell, R., Chopra, S., and LeCun, Y.
\newblock Dimensionality reduction by learning an invariant mapping.
\newblock In \emph{CVPR}, 2006.

\bibitem[He et~al.(2016)He, Zhang, Ren, and Sun]{DBLP:conf/cvpr/HeZRS16}
He, K., Zhang, X., Ren, S., and Sun, J.
\newblock Deep residual learning for image recognition.
\newblock In \emph{CVPR}, pp.\  770--778, 2016.

\bibitem[He et~al.(2019)He, Fan, Wu, Xie, and Girshick]{he2019momentum}
He, K., Fan, H., Wu, Y., Xie, S., and Girshick, R.
\newblock Momentum contrast for unsupervised visual representation learning.
\newblock \emph{arXiv:1911.05722}, 2019.

\bibitem[H{\'{e}}naff et~al.(2019)H{\'{e}}naff, Razavi, Doersch, Eslami, and
  van~den Oord]{CPC2}
H{\'{e}}naff, O.~J., Razavi, A., Doersch, C., Eslami, S. M.~A., and van~den
  Oord, A.
\newblock Data-efficient image recognition with contrastive predictive coding.
\newblock \emph{arXiv:1905.09272}, 2019.

\bibitem[Hermans et~al.(2017)Hermans, Beyer, and Leibe]{Hermans2017InDO}
Hermans, A., Beyer, L., and Leibe, B.
\newblock In defense of the triplet loss for person re-identification.
\newblock \emph{arXiv:1703.07737}, 2017.

\bibitem[Hjelm et~al.(2019)Hjelm, Fedorov, Lavoie{-}Marchildon, Grewal,
  Bachman, Trischler, and Bengio]{DIM}
Hjelm, R.~D., Fedorov, A., Lavoie{-}Marchildon, S., Grewal, K., Bachman, P.,
  Trischler, A., and Bengio, Y.
\newblock Learning deep representations by mutual information estimation and
  maximization.
\newblock In \emph{ICLR}, 2019.

\bibitem[Huang et~al.(2018)Huang, Yang, Lang, and Deng]{DBN}
Huang, L., Yang, D., Lang, B., and Deng, J.
\newblock Decorrelated batch normalization.
\newblock In \emph{CVPR}, 2018.

\bibitem[Ioffe \& Szegedy(2015)Ioffe and Szegedy]{DBLP:conf/icml/IoffeS15}
Ioffe, S. and Szegedy, C.
\newblock Batch normalization: Accelerating deep network training by reducing
  internal covariate shift.
\newblock In \emph{ICML}, 2015.

\bibitem[Ji et~al.(2019)Ji, Henriques, and Vedaldi]{IIC}
Ji, X., Henriques, J.~F., and Vedaldi, A.
\newblock Invariant information clustering for unsupervised image
  classification and segmentation.
\newblock In \emph{ICCV}, 2019.

\bibitem[Kingma \& Ba(2014)Kingma and Ba]{kingma2014adam}
Kingma, D.~P. and Ba, J.
\newblock Adam: A method for stochastic optimization.
\newblock \emph{arXiv:1412.6980}, 2014.

\bibitem[Krizhevsky \& Hinton(2009)Krizhevsky and Hinton]{CIFAR}
Krizhevsky, A. and Hinton, G.
\newblock Learning multiple layers of features from tiny images.
\newblock \emph{Technical Report}, 2009.

\bibitem[Le \& Yang(2015)Le and Yang]{tinyin}
Le, Y. and Yang, X.
\newblock Tiny imagenet visual recognition challenge.
\newblock 2015.

\bibitem[Mikolov et~al.(2013{\natexlab{a}})Mikolov, Chen, Corrado, and
  Dean]{DBLP:journals/corr/abs-1301-3781}
Mikolov, T., Chen, K., Corrado, G., and Dean, J.
\newblock Efficient estimation of word representations in vector space.
\newblock \emph{arXiv:1301.3781}, 2013{\natexlab{a}}.

\bibitem[Mikolov et~al.(2013{\natexlab{b}})Mikolov, Sutskever, Chen, Corrado,
  and Dean]{DBLP:conf/nips/MikolovSCCD13}
Mikolov, T., Sutskever, I., Chen, K., Corrado, G.~S., and Dean, J.
\newblock Distributed representations of words and phrases and their
  compositionality.
\newblock In \emph{NIPS}, 2013{\natexlab{b}}.

\bibitem[Misra \& van~der Maaten(2019)Misra and van~der
  Maaten]{misra2019selfsupervised}
Misra, I. and van~der Maaten, L.
\newblock Self-supervised learning of pretext-invariant representations.
\newblock \emph{arXiv:1912.01991}, 2019.

\bibitem[Misra et~al.(2016)Misra, Zitnick, and
  Hebert]{DBLP:conf/eccv/MisraZH16}
Misra, I., Zitnick, C.~L., and Hebert, M.
\newblock Shuffle and learn: Unsupervised learning using temporal order
  verification.
\newblock In \emph{ECCV}, 2016.

\bibitem[Noroozi \& Favaro(2016)Noroozi and Favaro]{DBLP:conf/eccv/NorooziF16}
Noroozi, M. and Favaro, P.
\newblock Unsupervised learning of visual representations by solving jigsaw
  puzzles.
\newblock In \emph{ECCV}, 2016.

\bibitem[Pathak et~al.(2016)Pathak, Kr{\"{a}}henb{\"{u}}hl, Donahue, Darrell,
  and Efros]{DBLP:journals/corr/PathakKDDE16}
Pathak, D., Kr{\"{a}}henb{\"{u}}hl, P., Donahue, J., Darrell, T., and Efros,
  A.~A.
\newblock Context encoders: Feature learning by inpainting.
\newblock \emph{CVPR}, 2016.

\bibitem[Richemond et~al.(2020)Richemond, Grill, Altché, Tallec, Strub, Brock,
  Smith, De, Pascanu, Piot, and Valko]{richemond2020byol}
Richemond, P.~H., Grill, J.-B., Altché, F., Tallec, C., Strub, F., Brock, A.,
  Smith, S., De, S., Pascanu, R., Piot, B., and Valko, M.
\newblock Byol works even without batch statistics.
\newblock \emph{arXiv:2010.10241}, 2020.

\bibitem[Schroff et~al.(2015)Schroff, Kalenichenko, and
  Philbin]{DBLP:conf/cvpr/SchroffKP15}
Schroff, F., Kalenichenko, D., and Philbin, J.
\newblock {FaceNet: A} unified embedding for face recognition and clustering.
\newblock In \emph{CVPR}, 2015.

\bibitem[Siarohin et~al.(2019)Siarohin, Sangineto, and
  Sebe]{siarohin2018whitening}
Siarohin, A., Sangineto, E., and Sebe, N.
\newblock Whitening and coloring transform for {GAN}s.
\newblock In \emph{International Conference on Learning Representations}, 2019.

\bibitem[Sohn(2016)]{DBLP:conf/nips/Sohn16}
Sohn, K.
\newblock Improved deep metric learning with multi-class n-pair loss objective.
\newblock In \emph{NIPS}, 2016.

\bibitem[Tian et~al.(2020{\natexlab{a}})Tian, Krishnan, and
  Isola]{DBLP:conf/eccv/TianKI20}
Tian, Y., Krishnan, D., and Isola, P.
\newblock Contrastive multiview coding.
\newblock In \emph{ECCV}, 2020{\natexlab{a}}.

\bibitem[Tian et~al.(2020{\natexlab{b}})Tian, Yu, Chen, and
  Ganguli]{tian2020understanding}
Tian, Y., Yu, L., Chen, X., and Ganguli, S.
\newblock Understanding self-supervised learning with dual deep networks.
\newblock \emph{arXiv:2010.00578}, 2020{\natexlab{b}}.

\bibitem[Tschannen et~al.(2019)Tschannen, Djolonga, Rubenstein, Gelly, and
  Lucic]{DBLP:journals/corr/abs-1907-13625}
Tschannen, M., Djolonga, J., Rubenstein, P.~K., Gelly, S., and Lucic, M.
\newblock On mutual information maximization for representation learning.
\newblock \emph{arXiv:1907.13625}, 2019.

\bibitem[van~den Oord et~al.(2018)van~den Oord, Li, and Vinyals]{CPC}
van~den Oord, A., Li, Y., and Vinyals, O.
\newblock Representation learning with contrastive predictive coding.
\newblock \emph{arXiv:1807.03748}, 2018.

\bibitem[Vincent et~al.(2008)Vincent, Larochelle, Bengio, and
  Manzagol]{denoising}
Vincent, P., Larochelle, H., Bengio, Y., and Manzagol, P.
\newblock Extracting and composing robust features with denoising autoencoders.
\newblock In \emph{ICML}, 2008.

\bibitem[Wang \& Isola(2020)Wang and Isola]{hypersphere}
Wang, T. and Isola, P.
\newblock Understanding contrastive representation learning through alignment
  and uniformity on the hypersphere.
\newblock In \emph{International Conference on Machine Learning}, 2020.

\bibitem[Wang \& Gupta(2015)Wang and Gupta]{DBLP:conf/iccv/WangG15}
Wang, X. and Gupta, A.
\newblock Unsupervised learning of visual representations using videos.
\newblock In \emph{ICCV}, 2015.

\bibitem[Wu et~al.(2018)Wu, Xiong, Yu, and Lin]{wu2018unsupervised}
Wu, Z., Xiong, Y., Yu, S., and Lin, D.
\newblock Unsupervised feature learning via non-parametric instance-level
  discrimination.
\newblock \emph{arXiv:1805.01978}, 2018.

\bibitem[Xu et~al.(2020)Xu, Chen, Moreno{-}Noguer, Jeni, and
  la~Torre]{DBLP:conf/eccv/XuCMJT20}
Xu, X., Chen, H., Moreno{-}Noguer, F., Jeni, L.~A., and la~Torre, F.~D.
\newblock {3D} human shape and pose from a single low-resolution image with
  self-supervised learning.
\newblock In \emph{ECCV}, 2020.

\bibitem[You et~al.(2017)You, Gitman, and Ginsburg]{you2017large}
You, Y., Gitman, I., and Ginsburg, B.
\newblock Large batch training of convolutional networks.
\newblock \emph{arXiv:1708.03888}, 2017.

\bibitem[Zhuang et~al.(2019)Zhuang, Zhai, and
  Yamins]{DBLP:journals/corr/abs-1903-12355}
Zhuang, C., Zhai, A.~L., and Yamins, D.
\newblock Local aggregation for unsupervised learning of visual embeddings.
\newblock \emph{arXiv:1903.12355}, 2019.

\end{thebibliography}
\bibliographystyle{icml2021}


\clearpage

\appendix

\section{Training Dynamics}
\label{train.plot}

\begin{figure*}[h]
\centering
\includegraphics[width=1\linewidth]{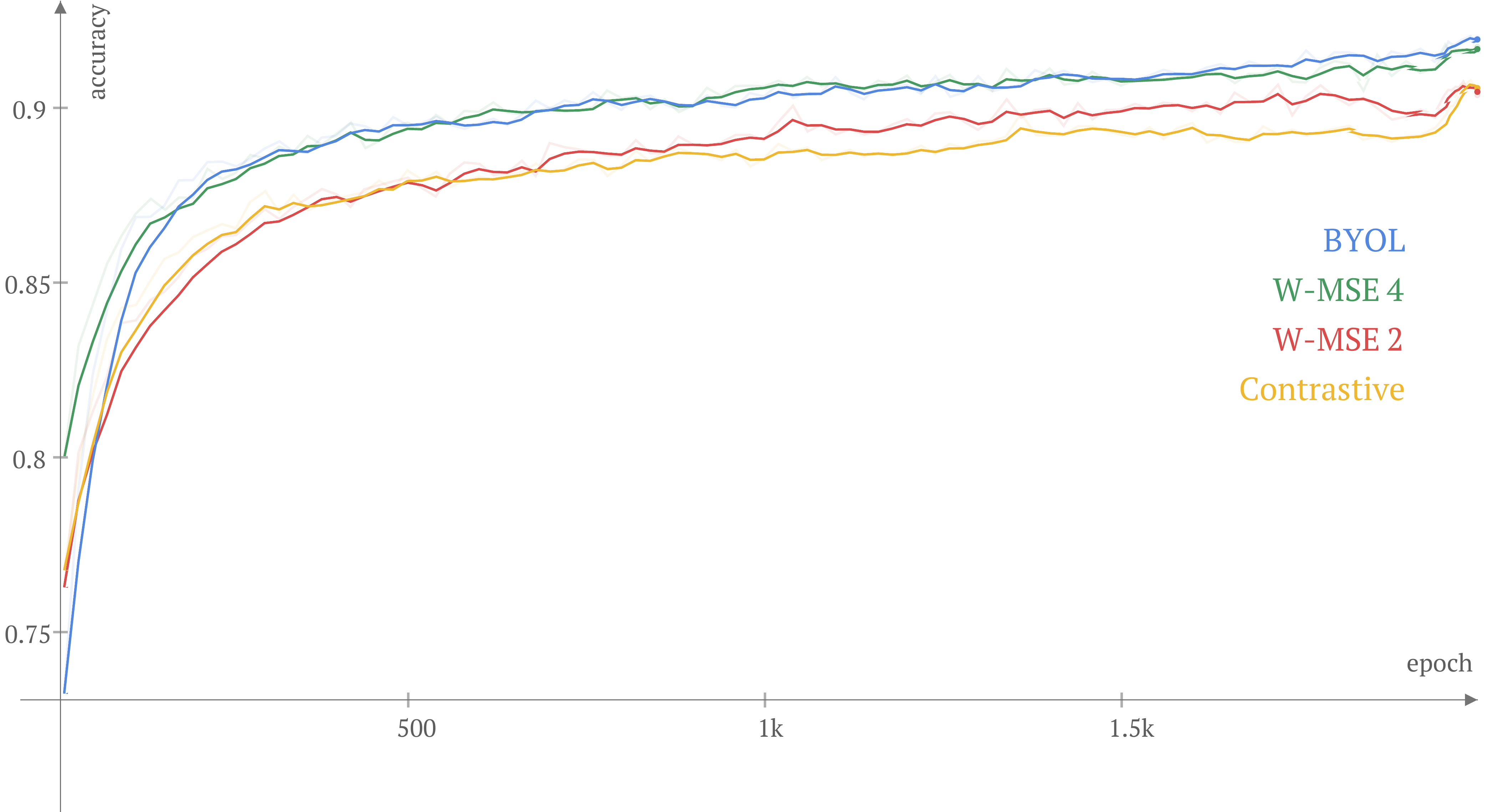}
\caption{Training dynamics on the STL-10 dataset (linear-classifier based evaluation).}
\label{fig.plot0}
\end{figure*}

\begin{figure*}[h]
\centering
\includegraphics[width=1\linewidth]{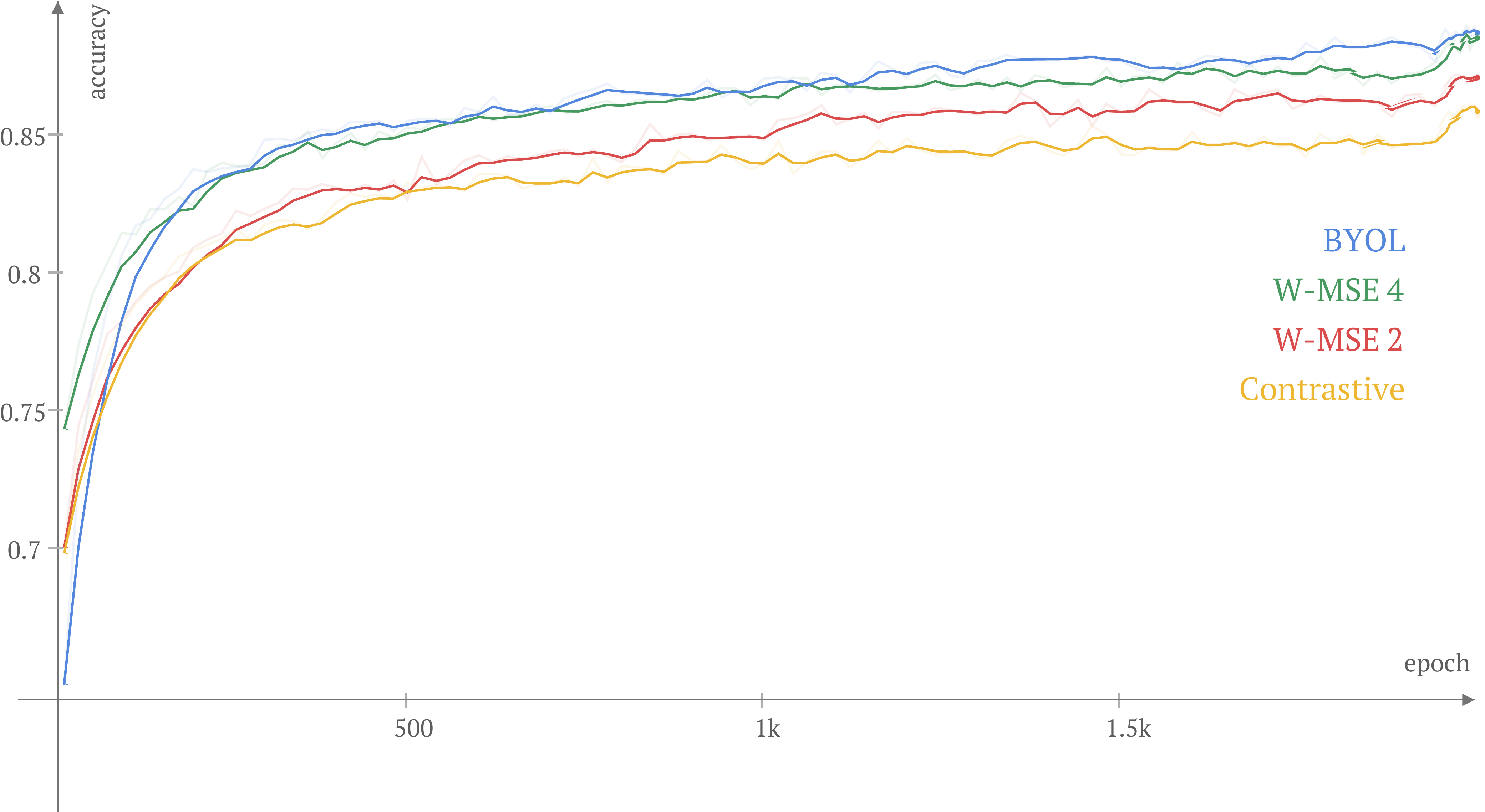}
\caption{Training dynamics on the STL-10 dataset (5-nn classifier based evaluation).}
\label{fig.plot1}
\end{figure*}

Fig.~\ref{fig.plot0} and \ref{fig.plot1} show the training dynamics for methods compared in Tab.~\ref{table.sota}. All the curves are smoothed with a 0.3 moving average for a better readability 
(curves before smoothing are shown semi-transparent).

\section{Cholesky Whitening and Backprogation}
\label{Cholesky}

We compute $W_V$ (Eq.~\ref{eq.sigma-V}) following \citep{siarohin2018whitening} and using the Cholesky decomposition. The Cholesky decomposition is based on the factorization of the covariance  symmetric matrix using two triangular matrices: $\Sigma_V = LL^\top$, where $L$ is a lower triangular matrix. Once we $L$ is computed, we compute the inverse of $L$, and we get: $W_V = L^{-1}$. Note that the Cholesky decomposition is fully diferentiable and it is implemented in all of the major frameworks, such as PyTorch and TensorFlow. However, for the sake of completeness, we provide below the gradient computation.

\subsection{Gradient Computation}

We provide here the equations for whitening differentiation as reported in \citep{siarohin2018whitening}. Let  $Z$ be the whitened version of the batch $V$, i.e., $Z = W_V (V - \boldsymbol{\mu}_V)$. The gradient $\frac{\partial L}{\partial V}$ can be computed by:
\begin{equation}
    \frac{\partial L}{\partial V} = \frac{2}{K-1} \frac{\partial L}{\partial \Sigma} V + W_V^T \frac{\partial L}{\partial Z}.
\end{equation}
\noindent where the partial derivative $\frac{\partial L}{\partial Z}$ is backpropogated, while  $\frac{\partial L}{\partial \Sigma}$ is computed as follows:
\begin{equation}
\label{eq.dl-ds}
    \frac{\partial L}{\partial \Sigma} = -\frac{1}{2} W_V^T \left(P \circ \frac{\partial L}{\partial W_V}  W_V^T + \left(P \circ \frac{\partial L}{\partial W_V} W_V^T\right)^T \right) W_V 
\end{equation}
\noindent 
In \eqref{eq.dl-ds}, $\circ$ is Hadamard product, while $\frac{\partial L}{\partial W_V}$ is:
\begin{equation}
    \frac{\partial L}{\partial W_V} =  \frac{\partial L}{\partial Z} V^T,
\end{equation}

and $P$ is:
\begin{equation*}
P = 
\begin{pmatrix} 
\frac{1}{2} & 0 & \cdots & 0 \\
1 &  \frac{1}{2} & \ddots & 0 \\ 
1 & \ddots & \ddots & 0 \\
1 & \cdots & 1 & \frac{1}{2}
\end{pmatrix}.
\end{equation*}

\section{Training time complexity}
\label{train.time}

Following \citep{siarohin2018whitening},  the complexity of the whitening transform is $O(k^3 + M k^2)$, where $k$ is the embedding dimension and $M$ is the  size of the sub-batch used in the batch slicing process. Since $k < M$ (see Sec.~\ref{Method-Whitening}), the whitening transform is $O(M k^2)$, which is basically equivalent to the forward pass of $M$ activations in a fully-connected layer connecting two layers of $k$ neurons each. In fact, the training  time is dominated by other architectural choices which are usually more computationally demanding than the loss computation. For instance, BYOL \citep{byol} needs 4 forward passes through 2 networks for each pair of positives. Hence, to evaluate the wall-clock time, we measure the time spent for one mini-batch iteration by all the methods compared in Tab.~\ref{table.sota}. We use
the STL-10 dataset, a ResNet-18 encoder and a server with one Nvidia Titan Xp GPU. Time of one iteration: Contrastive, 459ms; BYOL, 602ms; W-MSE 2, 478ms; W-MSE 4, 493ms. The 19ms difference between Contrastive  and W-MSE 2 is due to the whitening transform. Since the factual time is mostly related to the sample forward and backward passes, the $d (d-1)$ positive comparisons in Eq.~\ref{eq.WMSE}, do not significantly increase the wall-clock time of W-MSE 4 with respect to W-MSE 2.

\section{Euclidean distance}
\label{euclidean}

\begin{table}[th]
\begin{center}
\caption{Classification accuracy (top 1) using the Euclidean distance (unnormalized embeddings) on STL-10.\\}
\label{table.euclidean}
{\renewcommand{\arraystretch}{1.25}
\begin{tabular}{l|r r}
\toprule
Method & linear & 5-nn \\
\midrule
SimCLR (our repro.) & 78.00 & 71.07 \\
BYOL (our repro.)   & 80.83 & 74.94 \\
W-MSE 2         & 89.91 & 85.56 \\
W-MSE 4         & 90.40 & 87.09 \\
\bottomrule
\end{tabular}}
\end{center}
\end{table}

The cosine similarity is a crucial component in most of the current self-supervised learning approaches.
This is usually implemented with an $L_2$  normalization of the latent-space representations, which corresponds to projecting  the features on the surface of the unit  hypersphere. However, in our W-MSE, the whitening transform projects the representation onto a spherical distribution (intuitively, we can say on the whole unit hypersphere). Preserving the module of the features before the $L_2$ normalization  may be useful in some applications, e.g., clustering  the features after the projection head using a Gaussian mixture model.
 Tab.~\ref{table.euclidean} shows an experiment on the STL-10 dataset where we use unnormalized embeddings for all the methods (and $\tau = 1$ for the contrastive loss). Comparing Tab.~\ref{table.euclidean} with Tab.~\ref{table.sota}, the accuracy decrease of W-MSE is significantly smaller than in the other methods.

\clearpage

\end{document}